\documentclass[11pt,a4paper]{article}
\usepackage{times,latexsym}
\usepackage{url}
\usepackage[T1]{fontenc}

\usepackage{booktabs}
\usepackage{multicol}
\usepackage{multirow}
\usepackage{amsmath}
\usepackage{amssymb}
\usepackage{bbm}
\usepackage{graphicx}
\usepackage{enumitem}
\usepackage{subcaption}
\usepackage{mathtools}
\usepackage{fontawesome5}

\usepackage[acceptedWithA]{tacl2021v1}

\usepackage{xspace,mfirstuc,tabulary}

\newcommand{\update}[1]{#1}

\newif\iftaclinstructions
\taclinstructionsfalse %
\iftaclinstructions
\renewcommand{\confidential}{}
\renewcommand{\anonsubtext}{(No author info supplied here, for consistency with
TACL-submission anonymization requirements)}
\newcommand{\instr}
\fi

\iftaclpubformat %

\else

\fi

\title{Grammatical Error Correction Evaluation \\by Optimally Transporting Edit Representation}

\author{
  Takumi Goto\quad 
  Yusuke Sakai\quad 
  Taro Watanabe
  \ \\
  Nara Institute of Science and Technology, Japan
  \\
  \texttt{\{goto.takumi.gv7, sakai.yusuke.sr9, taro\}@is.naist.jp}
}

\date{}

\begin{document}
\maketitle
\begin{abstract}
Automatic evaluation in grammatical error correction (GEC) is crucial for selecting the best-performing systems. Currently, reference-based metrics are a popular choice, which basically measure the similarity between hypothesis and reference sentences. However, similarity measures based on embeddings, such as BERTScore, are often ineffective, since many words in the source sentences remain unchanged in both the hypothesis and the reference.
This study focuses on edits specifically designed for GEC, i.e., ERRANT, and computes similarity measured over the edits from the source sentence. To this end, we propose \emph{edit vector}, a representation for an edit, and introduce a new metric, UOT-ERRANT, which transports these edit vectors from hypothesis to reference using unbalanced optimal transport.
Experiments with SEEDA meta-evaluation show that UOT-ERRANT improves evaluation performance, particularly in the +Fluency domain where many edits occur. Moreover, our method is highly interpretable because the transport plan can be interpreted as a soft edit alignment, making UOT-ERRANT a useful metric for both system ranking and analyzing GEC systems. Our code is available from \faGithub \: \url{https://github.com/gotutiyan/uot-errant}.
\end{abstract}

\section{Introduction}
Grammatical Error Correction (GEC) task aims to automatically correct grammatical errors found in an input sentence regarding various grammatical items, such as verb tenses, noun numbers, and spelling mistakes. Many GEC systems have been \update{proposed~\cite{bryant-etal-2019-bea, omelianchuk-etal-2020-gector, rothe-etal-2021-simple, tarnavskyi-etal-2022-ensembling, katinskaia-yangarber-2024-gpt, omelianchuk-etal-2024-pillars}} with diversity in performance, and users need to select the system suitable for their purposes by referring to the results of automatic evaluation metrics. Considering that large language models are launched frequently and can also be used as correction systems, the role of automatic evaluation metrics is increasingly important for system selection.

A typical GEC metric is reference-based evaluation, which compares a hypothesis sentence output from a system with a reference sentence annotated by humans. The central idea of the reference-based evaluation is to quantify the similarity between the hypothesis and the reference. Considering the recent development of neural models, evaluation could be carried out based on the similarity of contextualized word representation, such as BERTScore~\cite{DBLP:conf/iclr/ZhangKWWA20}. However, since only tokens that are identified as errors are edited in GEC, many words remain unchanged in the hypothesis and reference. For this reason, the information of matching tokens becomes dominant in the sentence or word embeddings based metrics, which does not lead to a good evaluation metric~\cite{gong-etal-2022-revisiting}.

In this study, we solve this problem by introducing the similarity of edits. An edit represents the difference between an input sentence and its corrected version as a rewrite of fine-grained units, such as tokens or phrases, and can be automatically extracted using tools like ERRANT~\cite{felice-etal-2016-automatic, bryant-etal-2017-automatic}. By using edits as proxies, it is possible to compare the hypothesis and reference sentences not directly, but as changes from the input sentence. Inspired by methods that quantify the similarity of sentences or words as the similarity of their embeddings, this study proposes to measure similarity based on the embedding representation of edits.

\begin{figure*}[t]
\centering
\includegraphics[width=0.99\textwidth]{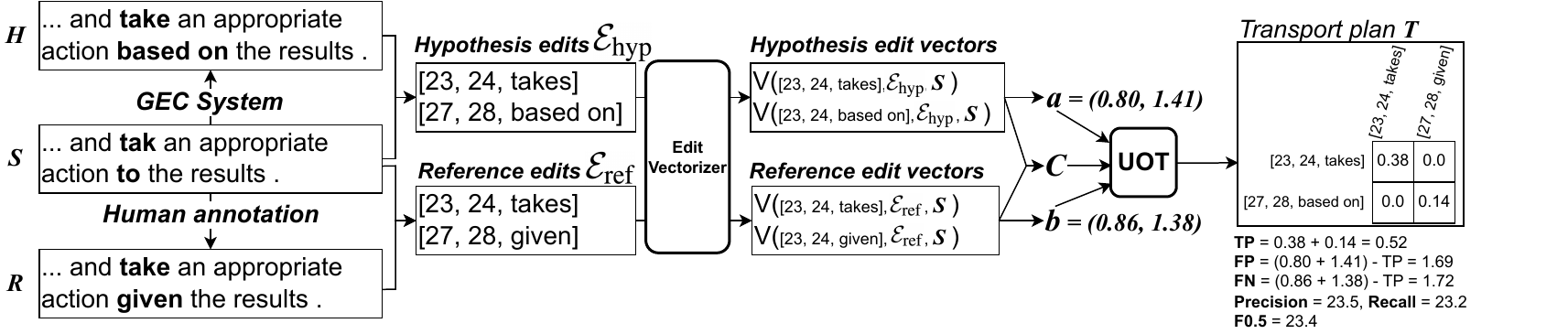}
\caption{
An overview of the proposed metric, UOT-ERRANT. Edits are extracted from the hypothesis and the reference, respectively, and converted into edit vectors. The optimal transport plan of edit vectors from the hypothesis to the reference vectors is then decomposed into a precision, recall, and $F_{0.5}$ score.
}
\label{fig:overview}
\end{figure*}

For this purpose, first, we propose an \emph{edit vector}, which is a vectorized representation of an edit measuring the impact on the sentence representation, defined as the difference vector between the embedding of the complete corrected sentence after applying all edits from an input sentence and a corrected sentence but reverting the corresponding single edit.
Next, we apply this to GEC evaluation by comparing the hypothesis and reference edit vectors using unbalanced optimal transport (UOT)~\cite{frogner2015learning}. In UOT, we transport the norm of edit vectors from the hypothesis to reference using the cost matrix defined over the $L^2$ distances of edits of two sentences. $L^2$ distance of two edits of a hypothesis and reference pair measures the closeness of edits, while the norm of an edit vector measures an impact on the corrected sentence of the single edit. The more transported values from the hypothesis to the reference imply more credit to the hypothesis.
Finally, by decomposing the transportation plan into precision, recall, and $F_{\beta}$, we adapt the UOT output to the general evaluation method of GEC. The overview shown in Figure~\ref{fig:overview} explains the process of the proposed metric: extraction of edits, vectorization, transportation by UOT, and decomposition of the transportation plan into evaluation values.
Our proposed metric, UOT-ERRANT, relies on edits from ERRANT,  but could be regarded as a soft-variant of edits when compared to M$^2$~\cite{dahlmeier-ng-2012-better} and ERRANT without strict surface form matching.

In our experiments, we mainly evaluate the performance of UOT-ERRANT on SEEDA~\cite{kobayashi-etal-2024-revisiting}, a representative meta-evaluation benchmark, and achieve results that surpassed conventional metrics, such as ERRANT~\cite{felice-etal-2016-automatic, bryant-etal-2017-automatic} and PT-ERRANT~\cite{gong-etal-2022-revisiting}, in correlation with human judgments. In particular, we demonstrated that evaluation performance is improved when requiring a smaller number of edits, and further gains are observed when many edits are involved by considering edits in contexts. We analyze the properties of the edit vector, focusing on error types~\cite{bryant-etal-2017-automatic}. We show that edits related to content words have larger norms and that they form clusters according to error type. These result indicates that the edit vector appropriately captures textual changes.

\section{Related Work}
\paragraph{Edit-level GEC Evaluation}
Edit-level metrics are the most representative ones for evaluating GEC systems, such as M$^2$~\cite{dahlmeier-ng-2012-better}, ERRANT~\cite{felice-etal-2016-automatic, bryant-etal-2017-automatic}, and PT-ERRANT~\cite{gong-etal-2022-revisiting}. Beyond word-wise edit metrics, chunk-wise units have also been proposed in prior studies~\cite{gotou-etal-2020-taking, ye-etal-2023-cleme}. These metrics can be interpreted as defining the similarity between a hypothesis sentence and a reference sentence as an edit or chunk matching rate. Essentially, an edit holds information about a span in the input sentence and its corresponding edited string. Existing metrics calculate the match between hypothesis edits and reference edits based on a complete match of this information. This evaluation method is considered a \emph{hard} alignment-based metric since it relies on a binary decision, i.e., a hypothesis edit is correct if it is included in the reference edits, and incorrect otherwise.
Hard alignment judges edits as incorrect solely based on differences in their surface forms, even when their impact for the semantic change is semantically accurate.

\paragraph{Quantifying Edits}
Several methods have been proposed to quantify the impact of edits for GEC evaluation. PT-ERRANT~\cite{gong-etal-2022-revisiting} employs BERTScore~\cite{DBLP:conf/iclr/ZhangKWWA20} to measure the different of a single edit,
which is used as an edit weight to improve the evaluation performance of edit-level metrics. IMPARA~\cite{maeda-etal-2022-impara} calculates the cosine similarity of the sentence representations between a corrected sentence and a sentence missing a single edit. It is used to generate pseudo-labels for training the quality estimation model, leading to high evaluation performance. 
Quantifying edits as a scalar in these studies limits their expression ability and prevents their use in similarity calculations. Indeed, PT-ERRANT relies on surface-level matches for edit alignment, with the scalar's role being solely for edit weighting.

\paragraph{NLP Applications of Optimal Transport}
Optimal transport~\cite{kantorovich1942translocation} has been applied to various NLP evaluation methods by quantifying the similarity between distributions. By treating words as discrete distributions based on their Word2Vec~\cite{mikolov2013efficient} or BERT-based word representations~\cite{devlin-etal-2019-bert}, the similarity between sentences is quantified based on a transport cost between words, known as Word Mover's Distance~\cite{kusner2015word}. The quantified similarity has been applied to tasks such as nearest neighbor search~\cite{kusner2015word, yurochkin2019hierarchical} and evaluation by measuring the similarity between hypotheses and \update{references~\cite{kilickaya-etal-2017-evaluating, chow-etal-2019-wmdo, zhao-etal-2019-moverscore, colombo-etal-2021-automatic}.}
Furthermore, the calculated transport plan can be interpreted as alignment between samples and this property has been used for various tasks such as word alignment~\cite{arase-etal-2023-unbalanced}, text matching~\cite{swanson-etal-2020-rationalizing}, and word-level translation quality estimation~\cite{kuroda-etal-2024-word}.

\section{Proposed Methods}
\subsection{Edit Vector}\label{subsec:edit-vector}
We propose an \emph{edit vector}, a vectorization method for an edit which takes into account the impact of the change toward sentence-level semantic representation.
Given an original sentence $S$ and its edited version $H$, we extract a set of edits $\mathcal{E} = \{e_i\}_{i=1}^{|\mathcal{E}|}$ using an edit extraction tool such as ERRANT~\cite{felice-etal-2016-automatic, bryant-etal-2017-automatic}. Next, for each individual edit $e \in \mathcal{E}$, the edit vector $V(e, \mathcal{E}, S)$ is calculated as the difference between the sentence representation of the complete edited sentence $S_{\mathcal{E}}$ and the corrupted variant $S_{\mathcal{E}\setminus \{e\}}$ that lacks the edit $e$ as follows:
\begin{equation}\label{eq:edit-vector}
V(e, \mathcal{E}, S) = \text{Enc}(S_{\mathcal{E}}) - \text{Enc}(S_{\mathcal{E}\setminus \{e\}}),
\end{equation}
where $S_{\mathcal{E}}$ is a sentence after applying editing set $\mathcal{E}$ to $S$.
$\text{Enc}(\cdot) \in \mathbb{R}^{d}$ is a sentence encoder that embeds a sentence into a $d$-dimensional representation, such as a mean pooling of BERT~\cite{devlin-etal-2019-bert} representation.

The edit vector represents how much the sentence semantic changes by applying an edit. The direction and length of the vector can be interpreted as quantifying how and to what extent an edit changes the sentence meaning, respectively. Because of this, we expect edits that bring about similar semantic changes to be encoded as similar vectors.
Compared to methods that quantify an edit as a scalar~\cite{gong-etal-2022-revisiting, maeda-etal-2022-impara}, edit vectors offer a richer representation of an edit's impact, enabling operations such as similarity calculations.

\subsection{Edit-level Similarity via Optimal Transport}\label{subsec:ot-for-edits}
\subsubsection{Discrete Optimal Transport}
Optimal Transport is the problem of finding a transport plan to move one distribution to another. The inputs are two discrete distributions $\boldsymbol{a} \in \mathbb{R}^n$ and $\boldsymbol{b} \in \mathbb{R}^m$, and a cost matrix $\mathbf{C} \in \mathbb{R}^{n\times m}_{+}$ between samples. The cost matrix ${C}_{ij}$ corresponds to the cost of transporting from $a_i$ to $b_j$. The output is the optimal transport plan $\mathbf{T} \in \mathbb{R}^{n\times m}_{+}$, where $T_{ij}$ denotes the amount of transport from sample $a_i$ to sample $b_j$. This plan minimizes the product of transport amounts and costs, $\sum_{ij} T_{ij} C_{ij}$. The plan $\mathbf{T}$ can also be interpreted as an alignment between samples, where samples with more transport are considered to be more strongly aligned.
To prevent the trivial solution where $\mathbf{T}$ becomes a zero matrix, which would minimize $\sum_{ij} T_{ij} C_{ij}$ without meaningful transport, additional constraints are generally added to ensure that meaningful transport occurs.

\paragraph{Balanced Optimal Transport (BOT)}
BOT~\cite{kantorovich1942translocation} adds the constraints that all of $\boldsymbol{a}$ must be transported, and $\boldsymbol{b}$ must receive all mass. Therefore, it is assumed that $\boldsymbol{a}$ and $\boldsymbol{b}$ have the same total mass, i.e., $\sum_{i=1}^{n} a_i = \sum_{i=1}^{m} b_i$. For example, probability distributions satisfy this condition. Formally, these constraints can be formulated as a combinatorial optimization problem:
\begin{equation}\label{eq:ot1}
\text{OT}(\boldsymbol{a}, \boldsymbol{b}, \mathbf{C}) = \underset{\mathbf{P} \in \pi(\boldsymbol{a}, \boldsymbol{b})}{\text{argmin}} \;\sum_{i, j}P_{ij} C_{ij}, 
\end{equation}
\begin{equation}\label{eq:ot2}
\begin{split}
\text{whe}&\text{re}\; \pi(\boldsymbol{a}, \boldsymbol{b}) =\\ &\{\mathbf{X} \in \mathbb{R}^{n\times m}_{+}:  \mathbf{X}\mathbbm{1}_m = \boldsymbol{a}, \mathbf{X}^\intercal \mathbbm{1}_n = \boldsymbol{b} \}.
\end{split}
\end{equation}
Here, $\mathbbm{1}_n$ is an $n$-dimensional vector with all elements are 1. This combinatorial optimization problem can be solved by an efficient solver~\cite{5459199}.

\paragraph{Unbalanced Optimal Transport (UOT)}
UOT~\cite{frogner2015learning} is a relaxed version of the BOT constraints. While BOT assumes that all samples between distributions are transported, this might not be appropriate for certain tasks. For example, in word alignment, it is not assured that all words may match, but spurious words in a sentence are left unaligned~\cite{arase-etal-2023-unbalanced}. However, the transport plan generated by BOT forces the transport of all samples, making it unable to handle such unaligned words. To resolve this issue, UOT replaces the $\mathbf{X}\mathbbm{1}_m = \boldsymbol{a}$ and $\mathbf{X}^\intercal \mathbbm{1}_n = \boldsymbol{b}$ terms in the BOT constraints with regularization terms. Formally, we seek a transport plan that minimizes the following objective function:
\begin{equation}\label{eq:uot-main}
\begin{split}
    \text{U}&\text{OT}(\boldsymbol{a}, \boldsymbol{b}, \mathbf{C}) = 
    \underset{\mathbf{P} \in \mathbb{R}^{n\times m}_{+}}{\text{argmin}}\; \sum_{i, j} C_{ij} P_{ij} + \epsilon H(\textbf{P}) \\
    & + \lambda_1 \text{KL}(\textbf{P}\mathbbm{1}_m, \boldsymbol{a}) + \lambda_2 \text{KL}(\textbf{P}^\intercal \mathbbm{1}_n, \boldsymbol{b}).
\end{split}
\end{equation}
Here, $H(\cdot)$ is an entropy regularization term, and $\text{KL}(\cdot)$ is the Kullback-Leibler divergence. Increasing the parameter $\epsilon$ leads to a more uniform transport between samples. Likewise, increasing $\lambda_1$ and $\lambda_2$ encourages transportation of all mass. This regularized optimization problem can be solved using the Sinkhorn algorithm~\cite{sinkhorn1964relationship, chizat2017scalingalgorithmsunbalancedtransport}.

\subsubsection{Connection between UOT and GEC}\label{subsec:connection}
We apply the UOT framework to measure edit-level similarity. The fundamental idea is to transport hypothesis edit vectors to reference edit vectors, and UOT is applied by treating $\boldsymbol{a}$ as the hypothesis edits, $\boldsymbol{b}$ as the reference edits, and $\mathbf{C}$ as the cost matrix based on the distances between the hypothesis and reference edit vectors. The more mass that can be transported, the better the hypothesis edits align with the reference edits, and thus the higher the similarity. The reason for using UOT constraints is that GEC systems often result in over-correction or under-correction, leading to null-alignments between edits. For instance, in over-correction, some of hypothesis edits become null-alignment, while in under-correction, some of reference edits become null-alignment. \update{Compared to existing studies that primarily focus on word transport~\cite{zhao-etal-2019-moverscore, colombo-etal-2021-automatic}, this study proposes edit transport.}

Formally, given a source sentence $S$, a hypothesis sentence $H$, and a reference sentence $R$, we extract a set of hypothesis edits $\mathcal{E}_{\text{hyp}}$ and a set of reference edits $\mathcal{E}_{\text{ref}}$. Tools such as ERRANT can be used to extract these edits, and we convert each of  extracted edits into an edit vector.
\begin{equation}\label{eq:edit-vectorize-hyp}
    \mathbf{V}^{\text{hyp}} = \{V(e, \mathcal{E}_{\text{hyp}}, S) | e \in \mathcal{E}_{\text{hyp}}\}
\end{equation}
\begin{equation}\label{eq:edit-vectorize-ref}
    \mathbf{V}^{\text{ref}} = \{V(e, \mathcal{E}_{\text{ref}}, S) | e \in \mathcal{E}_{\text{ref}}\}
\end{equation}
We define $\boldsymbol{a}$ and $\boldsymbol{b}$ as the norm of edit vectors $\mathbf{V}^{\text{hyp}}$ and $\mathbf{V}^{\text{ref}}$, respectively, and $\mathbf{C}$ as the Euclidean distance between edit vectors. 
As discussed in Section~\ref{subsec:edit-vector}, each edit vector represents the impact toward the change in sentence semantic with norm representing as its magnitude.
Thus, using the norm for $\boldsymbol{a}$ and $\boldsymbol{b}$ is regarded as an implicit weighting of edits. 
We assume that the Euclidean distance is an appropriate measure for the transport cost of edits, based on the intuition that similar edits are encoded into vectors with similar directions and norms.
Finally, we compute the optimal transport plan $\mathbf{T} = \text{UOT}(\boldsymbol{a}, \boldsymbol{b}, \mathbf{C})$. This plan represents edit-level alignment, and the total transported mass $\sum_{ij} T_{ij}$ is regarded as a similarity between two edit sets.

\subsection{Proposed Metric: UOT-ERRANT}\label{subsec:uot-errant}
As an application of the edit-level transport plan, we propose a new metric for GEC, namely \emph{UOT-ERRANT}. UOT-ERRANT calculates precision, recall, and $F_{\beta}$, which are typically used in GEC, by focusing on both the transported and non-transported mass from the transport plan. This quantifies the degree of over-correction and under-correction in addition to the simple similarity between the hypothesis and the reference. Figure~\ref{fig:overview} shows an example where both the hypothesis and the reference have two edits.

First, we decompose the transport plan $\mathbf{T}$ into True Positive (Score$_\text{TP}$), False Positive (Score$_\text{FP}$), and False Negative (Score$_\text{FN}$).
Score$_\text{TP}$ is the degree to which hypothesis edits match reference edits, calculated as the sum of transported mass:
\begin{equation}\label{eq:tp}
    \text{Score}_\text{TP} = \sum\nolimits_{i,j} T_{ij}.
\end{equation}
Score$_\text{FP}$ corresponds to the hypothesis edits that are considered incorrect, calculated as the sum of mass that could not be transported from the hypothesis:
\begin{equation}\label{eq:fp}
    \text{Score}_\text{FP} = \sum\nolimits_{i}a_i - \text{Score}_\text{TP}.
\end{equation}
Similarly, Score$_\text{FN}$ corresponds to reference edits that the GEC system missed, calculated as the sum of mass that was not transported to the reference edits:
\begin{equation}\label{eq:fn}
    \text{Score}_\text{FN} = \sum\nolimits_{j}b_j - \text{Score}_\text{TP}.
\end{equation}
Given Score$_\text{TP}$, Score$_\text{FP}$, and Score$_\text{FN}$, we calculate precision, recall, and $F_{\beta}$:
\begin{equation}
    \text{Precision} = \frac{\text{Score}_\text{TP}}{\text{Score}_\text{TP} + \text{Score}_\text{FP}},
\end{equation}
\begin{equation}
\text{Recall} = \frac{\text{Score}_\text{TP}}{\text{Score}_\text{TP} + \text{Score}_\text{FN}}.
\end{equation}
\begin{equation}
    F_{\beta} = \frac{\mathopen{}\left( 1 + \beta^{2} \mathclose{}\right) \text{Precision}\times \text{Recall}}{\beta^{2} \text{Precision} + \text{Recall}}.
\end{equation}
Following existing edit-level metrics~\cite{bryant-etal-2017-automatic, gong-etal-2022-revisiting}, we use $\beta=0.5$ as the final sentence-level score. When multiple references are available, the reference that yields the highest $F_{0.5}$ is selected for each input sentence, following existing reference-based metrics~\cite{bryant-etal-2017-automatic, gong-etal-2022-revisiting, koyama-etal-2024-n-gram}.

\section{Experiments}
\subsection{Meta-evaluation Settings}\label{subsec:meta-eval-setting}
\paragraph{Meta-Evaluation Dataset.}
\begin{figure}[t]
\centering
\includegraphics[width=0.48\textwidth]{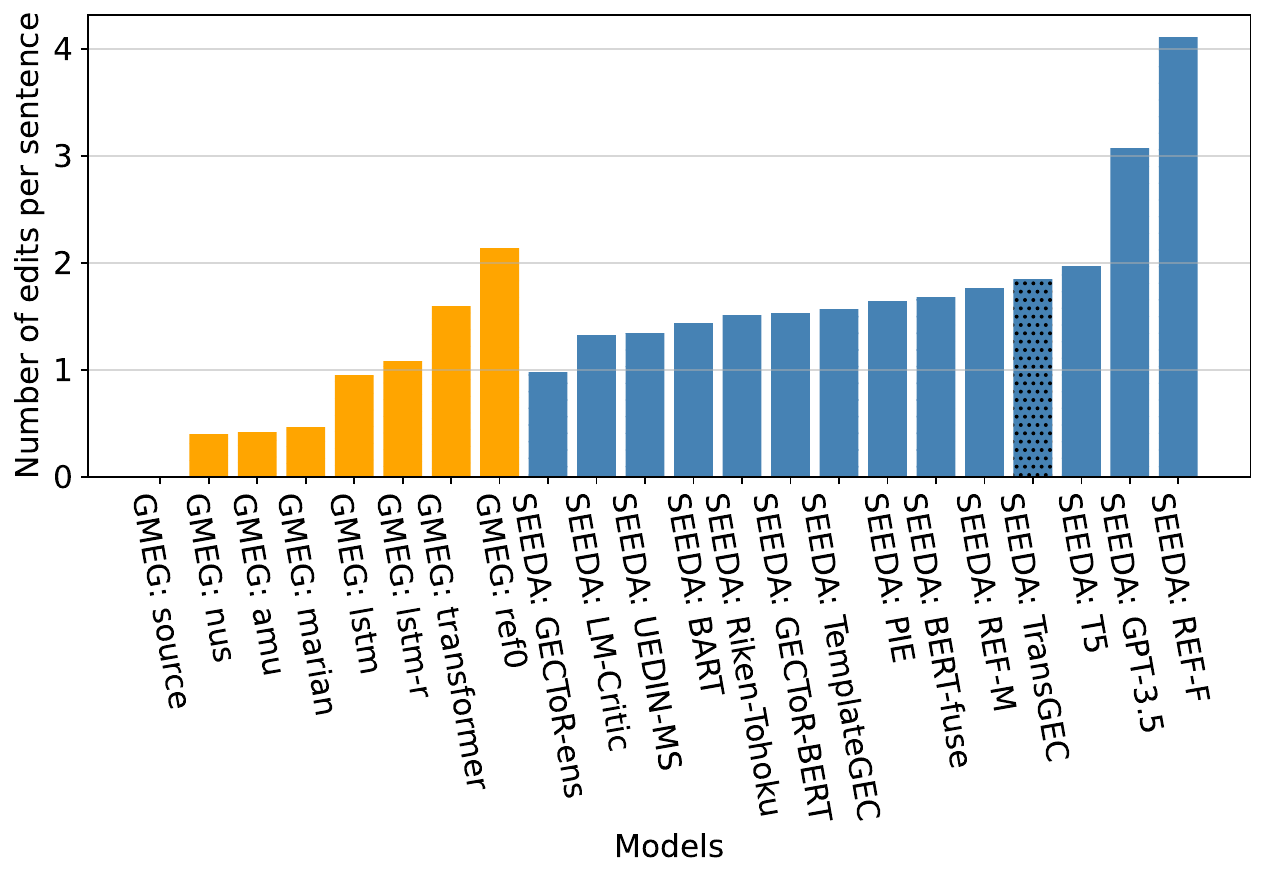}
\caption{\update{System-wise average number of edits per sentence in both Wiki domain of GMEG-Data (Orange) and SEEDA (Blue). For each dataset, the bars are sorted by the heights.}}
\label{fig:density}
\end{figure}

Our proposed method is evaluated using the meta-evaluation benchmark SEEDA~\cite{kobayashi-etal-2024-revisiting} \update{and GMEG-Data~\cite{napoles-etal-2019-enabling}}. SEEDA is a dataset that includes the outputs of 14 systems, including newer models such as GPT-3.5~\cite{brown2020language}, and their human evaluation rankings. It supports two types of domains according to the correction tendencies of the systems: the Base setting that ranks 12 systems excluding GPT-3.5 outputs and fluent reference sentences, and the +Fluency setting that ranks all 14 systems including them.
There exist two variants for each setting: SEEDA-E, based on edit-level human evaluation, and SEEDA-S, based on sentence-level human evaluation. \citet{kobayashi-etal-2024-revisiting} reported the importance of matching the granularity of evaluation between human and automatic evaluations. Since UOT-ERRANT is an edit-level metric, we evaluate it on SEEDA-E.
\update{GMEG-Data is a meta-evaluation dataset that incorporates writing by native speakers. While SEEDA and GJG15 target writing by language learners, GMEG-Data enables meta-evaluation in scenarios with fewer errors. In this study, we use the Wiki domain, which covers the formal style written by native speakers. The outputs to be evaluated consist of eight systems: the source, human correction results (reference), and six GEC systems. In the original paper, the human correction results were assessed by sampling from multiple references; however, because the results of that sampling have not been released, we cannot reproduce it. Given that four references are available in GMEG-Data, we treat the first reference as a reference system in our experiments. As shown in Figure~\ref{fig:density}, the GEC systems in the Wiki domain made fewer edits per sentence than those for SEEDA, facilitating meta-evaluation in situations with lower error density.}

\paragraph{Computation of System-level Scores.}

Human evaluation calculates system-level scores by aggregating sentence-level scores. Specifically, it converts sentence-level scores into pairwise comparison results by comparing systems against each other, then transforms them into system-level scores using rating algorithms such as TrueSkill~\cite{NIPS2006_f44ee263} or Expected Wins~\cite{bojar-etal-2013-findings}. In practice, SEEDA uses TrueSkill-based scores as their final scores.
Our implementation uses \textsc{gec-metrics}~\cite{goto-etal-2025-gec}\footnote{\url{https://pypi.org/project/gec-metrics/}} with each system's default settings.
\citet{goto-etal-2025-rethinking} pointed out that aggregation methods in previous automatic evaluations differed from human evaluations and reported that aligning them improved evaluation performance. Following this insight, our experiments adopt TrueSkill-based aggregation for all metrics.

\paragraph{Reference Set.}
While SEEDA provides source, hypotheses, and human evaluations for the hypotheses, it does not provide references.
To assess the robustness of the metrics, we conduct meta-evaluation with diverse reference sets, following \citet{ye-etal-2023-cleme}. \textbf{Official} refers to the two official references used in the CoNLL-2014 shared task~\cite{ng-etal-2014-conll}. \textbf{10 Refs} consists of 10 unofficial references annotated by \citet{bryant-ng-2015-far}. \textbf{E-Minimal}, \textbf{NE-Minimal}, \textbf{E-Fluency}, and \textbf{NE-Fluency} are annotated by \citet{sakaguchi-etal-2016-reassessing}. These are combinations of the annotator's expertise, expert (E-) and non-expert (NE-), and domains regarding the purpose of GEC, minimal edit (-Minimal), and fluency edit (-Fluency). For the SEEDA-E Base, we use four types: Official, 10refs, E-Minimal, and NE-Minimal, and for +Fluency, we use E-Fluency and NE-Fluency. Note that, since one of the E-Minimal and one of the E-Fluency are already used as SEEDA systems, they are excluded and used as a single reference, respectively. \update{For GMEG-Data, although it provides four references, we use the second, third, and fourth references as a reference set because we use the first reference as a GEC system to be evaluated.}

\begin{table*}[t]
    \small
    \centering
    \setlength{\tabcolsep}{2.2pt}
    \resizebox{\linewidth}{!}{%
    \begin{tabular}{@{}l|cc|cc|cc|cc|cc|cc|cc|c@{}}
    \toprule
    & \multicolumn{8}{c|}{\textbf{↑ SEEDA-E Base Setting}} & \multicolumn{4}{c|}{\textbf{↑ SEEDA-E +Fluency Setting}} & \multicolumn{2}{c|}{\update{\textbf{↑ GMEG-Data}}} & \\
     \cmidrule(lr){2-9} \cmidrule(lr){10-13} \cmidrule(lr){14-15}
    & \multicolumn{2}{c}{Official} & \multicolumn{2}{c}{10 Refs} & \multicolumn{2}{c}{E-Minimal} & \multicolumn{2}{c|}{NE-Minimal} & \multicolumn{2}{c}{E-Fluency} & \multicolumn{2}{c|}{NE-Fluency} & \multicolumn{2}{c}{} &\\
        \cmidrule(lr){2-3} \cmidrule(lr){4-5} \cmidrule(lr){6-7} \cmidrule(lr){8-9} \cmidrule(lr){10-11} \cmidrule(lr){12-13} \cmidrule(lr){14-15} 
        Metrics & $r$ & $\rho$ & $r$ & $\rho$ & $r$ & $\rho$ &  $r$ & $\rho$ &  $r$ & $\rho$ & $r$ & $\rho$ &  $r$ & $\rho$ &  ↓ Avg. Rank \\ 
    \midrule
\multicolumn{16}{l}{$n$-gram level and reference-based metrics} \\
\midrule

GLEU & .909 & \underline{.965} & .949 & .958 & .848 & .916 & \underline{.808} & \underline{.895} & .278 & .600 & \textbf{.781} & \textbf{.921} & .785 & .619  & 3.214\\
GREEN & .912 & \underline{.965} & .910 & .979 & .858 & \underline{.930} & .700 & .825 & \textbf{.547} & \textbf{.802} & \underline{.745} & \underline{.908} & \textbf{.871} & \textbf{.786}  & \underline{2.786}\\
\midrule
\multicolumn{16}{l}{Edit-level and reference-based metrics} \\
\midrule
ERRANT & .881 & .895 & \underline{.952} & .951 & .864 & .804 & .740 & .720 & -.005 & .424 & .114 & .508 & .780 & .667 & 4.857\\
PT-ERRANT & \underline{.924} & .951 & \textbf{.957} & \underline{.986} & .894 & .888 & .802 & .832 & .005 & .310 & .276 & .578 & \underline{.830} & \underline{.690} & 3.286\\
CLEME  & .910 & .930 & .932 & .937 & \textbf{.910} & \textbf{.965} & .716 & .706 & -.043 & .297 & .473 & .653 & .825 & .667 & 4.286\\
UOT-ERRANT & \textbf{.950} & \textbf{.979} & .942 & \textbf{.993} & \underline{.901} & .909 & \textbf{.915} & \textbf{.930} & \underline{.445} & \underline{.684} & .705 & .851 & .699 & \underline{.690} & \textbf{2.357}\\
\midrule
\multicolumn{16}{l}{Sentence-level and reference-free metrics} \\
\midrule
SOME   & \multicolumn{8}{c|}{.893 / .944} & \multicolumn{4}{c|}{.965 / .965} & .818 & .548 & - \\
Scribendi & \multicolumn{8}{c|}{.837 / .888} & \multicolumn{4}{c|}{.826 / .912} & .822 & .571 & - \\
IMPARA & \multicolumn{8}{c|}{.901 / .944} & \multicolumn{4}{c|}{.969 / .965} & .868 & .619  & - \\
Qwen3-8B-S & \multicolumn{8}{c|}{.886 / .965} & \multicolumn{4}{c|}{.418 / .939} & .869 & .976 & - \\
GPT-4-E$^{\spadesuit}$ & \multicolumn{8}{c|}{.905 / .986} & \multicolumn{4}{c|}{.848 / .987} & N/A & N/A & - \\
    \bottomrule
    \end{tabular}
    }
    \caption{System-level meta-evaluation results on SEEDA-E, with various reference sets. $r$ is Pearson correlation and $\rho$ is Spearman correlation. \textbf{Bold} represents the highest value in each column, and \underline{underline} represents the second highest. ``Avg. Rank'' refers to the average rank when the results in each column are converted into rankings, indicating the robustness of reference-based metrics to the reference. \update{The symbol ${\spadesuit}$ denotes that we cite the values reported in the original paper~\cite{kobayashi-etal-2024-large}.}
}
    \label{tab:system-results}
\end{table*}

\paragraph{UOT-ERRANT Settings.}\label{subsec:uot-errant-setting}
We use mean pooling representation of ELECTRA base~\footnote{\url{google/electra-base-discriminator}}~\cite{DBLP:conf/iclr/ClarkLLM20} for Enc$(\cdot)$ in Equation~\ref{eq:edit-vector}.
We use the stabilized algorithm~\cite{schmitzer2019stabilized} implemented in the POT library~\cite{flamary2021pot, flamary2024pot} to solve the UOT problem.
The calculation of Equation~\ref{eq:uot-main} requires three hyperparameters: $\epsilon$, $\lambda_1$, and $\lambda_2$. Since direct parameters tuning on SEEDA is not possible due to test leakage, we tune these parameters using GJG15~\cite{grundkiewicz-etal-2015-human} as development data. GJG15 comprises the 12 system outputs submitted to CoNLL-2014 shared task~\cite{ng-etal-2014-conll} and their human evaluation. As there is no need to uniformize the transport plan, we fix the weight of the entropy regularization term $\epsilon$ to 0.1, following \citet{arase-etal-2023-unbalanced}. We assume that $\lambda_1 = \lambda_2$ and vary them in the range [0.02, 1.00] with a step of 0.02. We use the parameters that yield the highest Pearson correlation with human evaluations in GJG15. When the Pearson correlation is the same, we use parameters with the higher Spearman correlation.
Finally, we use $\lambda_1 = \lambda_2 = 0.1$.

\subsection{Existing Metrics to be Compared}
\paragraph{ERRANT~\cite{felice-etal-2016-automatic, bryant-etal-2017-automatic}} is a representative edit-level metric in GEC. Similar to Section~\ref{subsec:connection}, after extracting the edits $\mathcal{E}_{\text{hyp}}$ and $\mathcal{E}_{\text{ref}}$, we calculate precision, recall, and $F_{0.5}$ from their overlap, $\mathcal{E}_{\text{overlap}} = \mathcal{E}_{\text{hyp}} \cap \mathcal{E}_{\text{ref}}$:
\begin{equation}
\label{eq:errant}
    \text{Prec.} = \frac{
    \sum_{e \in \mathcal{E}_{\text{overlap}}} w_e
    }{
    \sum_{e \in \mathcal{E}_{\text{hyp}}} w_e
    },
    \text{Recall} = \frac{
    \sum_{e \in \mathcal{E}_{\text{overlap}}} w_e
    }{
    \sum_{e \in \mathcal{E}_{\text{ref}}} w_e
    }.
\end{equation}
The $w_e$ is the weight of an edit $e$, and ERRANT always uses $w_e = 1$.

\paragraph{PT-ERRANT~\cite{gong-etal-2022-revisiting}} is based on the same formulation as Equation~\ref{eq:errant}, but calculates $w_e$ using BERTScore~\cite{DBLP:conf/iclr/ZhangKWWA20}. Specifically, it uses the absolute value of the difference in scores when applying the edit to the input text as the weight: $w_e = |\text{BERTScore}(S, R) - \text{BERTScore}(S_{\{e\}}, R)|$. We use \texttt{bert-based-uncased}~\footnote{\url{google-bert/bert-base-uncased}} for calculating BERTScore.

\paragraph{GLEU~\cite{napoles-etal-2015-ground, napoles2016gleutuning} and GREEN~\cite{koyama-etal-2024-n-gram}} are reference-based $n$-gram metrics. GLEU is a precision-based metric, while GREEN enables the calculation of $F_{\beta}$. For both metrics, we use $n$ ranging from 1 to 4. Following \citet{koyama-etal-2024-n-gram}, we use $F_{2.0}$ in GREEN.

\paragraph{CLEME~\cite{ye-etal-2023-cleme}} compares the chunk sequences constructed from the edits between the hypothesis and reference. We use CLEME-independent, and the scale factors and thresholds for TP, FP, and FN follow~\citet{ye-etal-2023-cleme}.

\paragraph{SOME~\cite{yoshimura-etal-2020-reference}} is a reference-free metric based on a weighted sum of scores for grammaticality, fluency, and meaning preservation. We use the official pre-trained weights~\footnote{\url{https://github.com/kokeman/SOME}}, and 0.55, 0.43, and 0.02 as the weights for grammaticality, fluency, and meaning preservation, respectively. These are the weights tuned for sentence-level evaluation by~\citet{yoshimura-etal-2020-reference}.

\paragraph{Scribendi~\cite{islam-magnani-2021-end}} evaluates hypotheses by confirming that the perplexity calculated by the language model has decreased and that surface consistency has been maintained. GPT-2~\cite{radford2019language} is used as the language model, and 0.8 is used as the threshold for surface consistency.

\paragraph{IMPARA~\cite{maeda-etal-2022-impara}} uses a similarity model to confirm that the meaning has been preserved by the correction, and then calculates the score using a quality estimation model. We use \texttt{bert-base-cased} as the similarity model, the unofficial quality estimation model published by~\citet{goto-etal-2025-gec}~\footnote{\url{https://huggingface.co/gotutiyan/IMPARA-QE}. Note that no official pre-trained models are available.}, and a similarity threshold of 0.9.

\paragraph{LLM-S and LLM-E~\cite{kobayashi-etal-2024-large}}
are an LLM-as-a-judge-based metric that takes multiple hypotheses and estimates a 5-level score for each of the hypotheses. 
\update{The difference between LLM-S and LLM-E lies in whether the hypothesis is input as a sentence or converted into an editing sequence before input.
\citet{goto-etal-2025-gec} mentions that they could not fully reproduce the results reported in the original paper, particularly for LLM-E. Thus, we cite the values of GPT-4-E reported in \citet{kobayashi-etal-2024-large}, when using GPT-4~\cite{openai2024gpt4technicalreport} as a LLM, to avoid underestimating this metric. Furthermore, since \citet{kobayashi-etal-2024-large} did not report results for GMEG-Data, we also report our LLM-S experimental results based on Qwen3-8B~\cite{yang2025qwen3technicalreport}, denoted as Qwen3-8B-S.
}

\subsection{Experimental Results}\label{subsec:results}
Table~\ref{tab:system-results} shows the correlations with human evaluation in SEEDA-E. The proposed method achieved superior performance over existing metrics across various reference sets. In Base, the proposed method often ranks first, and in +Fluency, it achieved the highest correlation among edit-level metrics. +Fluency evaluates performance in situations where more edits occur, leading to a wider variety of edit patterns compared to Base. In such scenarios, it is challenging to capture diverse correct edits with a small number of references~\cite{choshen-abend-2018-inherent, bryant-ng-2015-far}. Our method solves this problem by considering edit similarity via edit vector. ``Avg. Rank'' in Table~\ref{tab:system-results} shows the average rank of the correlations in each column for reference-based metrics. Our method achieved the highest average rank among the reference-based metrics, indicating that it enables robust evaluation across diverse domains and reference sets.

\begin{figure}[t]
\centering
\includegraphics[width=0.45\textwidth]{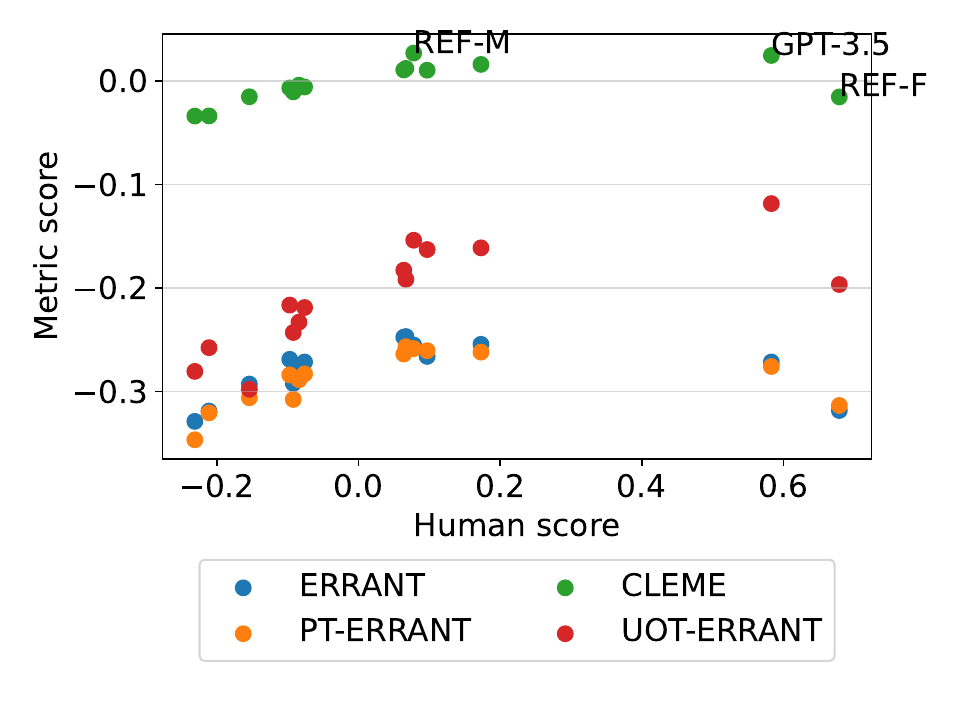}
\caption{
Scatter plot for 14 systems on SEEDA-E +Fluency. The $x$-axis and $y$-axis represent human and metric scores, respectively.
}
\label{fig:scatter}
\end{figure}

\begin{figure}[t]
    \centering
    \includegraphics[width=0.45\textwidth]{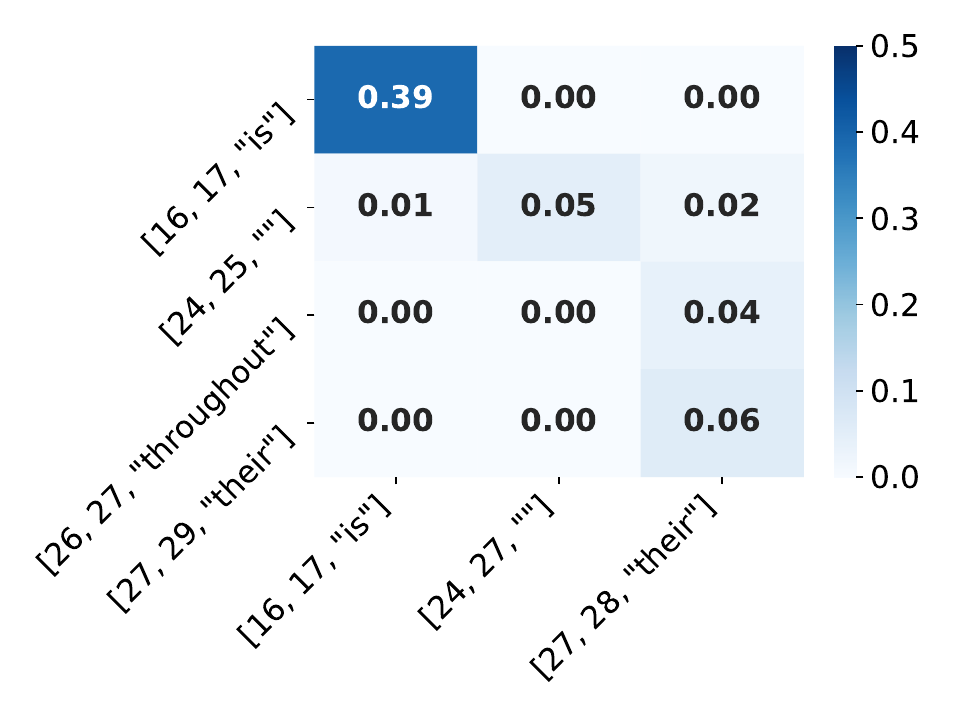}
        \caption{A case study: Alignment between four hypothesis edits ($y$-axis) and three reference edits ($x$-axis). The actual sentences are as follows:
        \\\textbf{Source}: ``It is still early for parents to decide whether they can foster a new life that \underline{are} not able to work and may suffer \underline{the pain in the entire} life .''
        \\\textbf{Reference}: ``$\dots$ new life that \underline{is} not able to work and may
suffer \underline{their} entire life .''
        \\\textbf{Hypothesis}: ``$\dots$ new life that \underline{is} not able to work and
may suffer pain \underline{throughout} \underline{their} life .''}
    \label{fig:case-study-391}
\end{figure}

\update{In GMEG-Data, UOT-ERRANT showed the second-highest Spearman correlation among reference-based metrics, but the lowest Pearson correlation. This result is likely due to differences in correction tendencies within the dataset. GMEG-Data contains more punctuation edits compared to SEEDA. Punctuation edits account for 32.1\% (2,241 / 6,976) of the total edits across all systems, which is in contrast to SEEDA's 5.9\% (599 / 10,062). Since edit vectors are intended to capture the semantic similarity of edits, they are less effective for punctuation edits, making them unsuitable for evaluating GMEG-Data. Therefore, the proposed method is an effective metric in situations where the surface forms of edits are diverse, such as in SEEDA's +Fluency.}

To investigate the factors behind the improved correlation in +Fluency, we present a scatter plot in Figure~\ref{fig:scatter}. The $x$-axis and $y$-axis represent the human evaluation scores and metric scores, respectively, for 14 systems with the representative system labels of \emph{REF-M}, \emph{GPT-3.5}, and \emph{REF-F}. REF-M includes a minimal correction, while GPT-3.5 and REF-F include relatively more edits.
The ideal plot would show a direct alignment on the diagonal line from the bottom-left to the top-right.
The difference between SEEDA's Base and +Fluency lies in whether GPT-3.5 and REF-F are included in the system set. Therefore, the performance improvement in +Fluency should be due to improved evaluation of these two systems.
The results show that UOT-ERRANT is able to assign relatively high scores to the two systems. ERRANT, PT-ERRANT, and CLEME assign higher scores to REF-M than to GPT-3.5, leading to lower correlation.

UOT-ERRANT also offers high explainability by visualizing the alignment between edits. Figure~\ref{fig:case-study-391} shows a transport plan using an actual example. This transport plan can be interpreted as an alignment that explains how the hypothesis edits match the reference edits.
In this example, existing metrics based on strict matching, such as ERRANT, only recognize the hypothesis edit [16, 17, ``is''] as correct, while evaluating the other three hypothesis edits as incorrect. In contrast, UOT-ERRANT assigns a partial score by aligning the hypothesis edit [24, 25, ``''] with the reference edit [24, 27, ``''] and [27, 29, ``their''] with [27, 28, ``their'']. Intuitively, these edits are not completely incorrect, and UOT-ERRANT metric captures this intuition by flexible edit-level evaluation based on the transport plan.

Although Table~\ref{tab:system-results} confirms that UOT-ERRANT improves correlation, especially under +Fluency, its correlation remains lower compared to $n$-gram-level and sentence-level metrics. However, the actual role of a metric is not only to rank systems, but also to analyze weaknesses and strengths of GEC systems.
Correlations with human evaluation commonly used in meta-evaluation reflect only ranking performance and are not suitable for measuring analytical capabilities.
Such analytical functionality is limited to edit-level metrics, and UOT-ERRANT achieves the highest evaluation performance among them.
In this sense, UOT-ERRANT offers the advantage of enabling analysis based on precision, recall, $F_{\beta}$, and edit alignments, which is shown in Figure~\ref{fig:case-study-391}.

\begin{table}[t]
    \centering
    \setlength{\tabcolsep}{4pt}
    \small
    \begin{tabular}{p{3cm}|cc|cc}
    \toprule
    & \multicolumn{2}{c|}{Base} & \multicolumn{2}{c}{+Fluency} \\
    & \multicolumn{2}{c|}{Official} & \multicolumn{2}{c}{NE-Fluency} \\
    \cmidrule(lr){2-3} \cmidrule(lr){4-5}
    & $r$ & $\rho$ & $r$ & $\rho$ \\
    \midrule
    \multicolumn{5}{l}{\textbf{Default settings} (same as Table~\ref{tab:system-results})} \\
    \midrule
    UOT-ERRANT & .950 & .979 & .705 & .851 \\
    \midrule
    \multicolumn{5}{l}{\textbf{Vectorization} (Default: By removing an edit))} \\
    \midrule
    By adding an edit &.956 & .979 & .430 & .657\\
    \midrule
    \multicolumn{5}{l}{\textbf{Mass} (Default: $L^2$ norm)} \\
    \midrule
    Uniform & .932 & .944 & .678 & .815 \\
    \midrule
    \multicolumn{5}{l}{\textbf{Cost} (Default: Euclidean distance)} \\
    \midrule
    Cosine similarity & .920 & .937 & .453 & .714 \\
    \midrule
    \multicolumn{5}{l}{\textbf{$\text{Enc}(\cdot)$} (Default: ELECTRA-base)} \\
    \midrule
    RoBERTa-base &.959 & .916 & .315 & .543\\
    LaBSE & .893 & .930 & .879 & .960\\
    \bottomrule
    \end{tabular}
    \caption{Ablation study on SEEDA-E. The \emph{UOT-ERRANT} on the top line indicates the results with our default settings, which is the same results as Table~\ref{tab:system-results}. The following lines show results when replacing the default setting with others.}
    \label{tab:ablation}
\end{table}

\subsection{Ablation Studies}\label{subsec:ablation}
We conduct ablation studies to verify how the hyperparameters of UOT-ERRANT improve GEC evaluation. We measure the meta-evaluation performance when varying hyperparameters under two configurations: Base with Official references, and +Fluency with NE-Fluency references. 
All results regarding UOT is based on the experimental setting described in Section~\ref{subsec:uot-errant-setting}.

\paragraph{Vectorization}
Edit vectors quantify the semantic impact of edits by removing them from a corrected sentence, but they can also be quantified by adding edits to the source, similar to PT-ERRANT's weight calculation. The difference between the two is whether the vectors are contextualized by other edits within the same sentence. In the example in Figure~\ref{fig:overview}, the edit [23, 24, ``takes''] is in both the hypothesis and reference edits. In adding an edit, it is quantified as the same vector in both edit sets, but in removing edit, surrounding edits such as [27, 28, ``given''] in the reference edits also influence the vector.
The results in \emph{By adding an edit} of Table~\ref{tab:ablation} show that removing an edit is better than adding it for vectorization on +Fluency. This suggests that the edit vectors contexualized by surrounding edits accurately quantifies the edit's impact for semantic changes. 

\paragraph{Mass}
We use uniform weights instead of the default $L^2$ norm, in order to measure the contribution of our proposed edit weighting as mentioned in Section~\ref{subsec:connection}. The correlation values in \emph{Uniform} in Table~\ref{tab:ablation} show that using the $L^2$ norm is better than the uniform. This result is consistent with previous research that weighting edits improves correlation, as reported in PT-ERRANT~\cite{gong-etal-2022-revisiting} and CLEME~\cite{ye-etal-2023-cleme}.

\paragraph{Cost}
The default setting is Euclidean distance, but we use cosine similarity instead to quantify the vector's length in the cost. The correlation values in \emph{Cosine similarity} in Table~\ref{tab:ablation} shows that Euclidean distance is better than cosine similarity, indicating that the vector's length is important for the cost definition.

\paragraph{Sentence Encoder}
We use 
RoBERTa-base~\cite{liu2019robertarobustlyoptimizedbert} and LaBSE~\cite{feng-etal-2022-language} instead of ELECTRA-base as a sentence encoder, Enc($\cdot$) in Equation~\ref{eq:edit-vector}. We use the mean pooling representation for RoBERTa, and \texttt{[CLS]} representation for LaBSE, considering pre-trained setting. The results are shown in the most bottom block in Table~\ref{tab:ablation}. 
Drop for RoBERTa-base implies the importance of the strong signals for each edit in the encoder.
The gains in LaBSE in +Fluency reflect the importance of the interactions of edits in the corrected sentence, but the drops in Base shows its weakness to quantify a single edit.

\section{Discussion}

\subsection{Characteristics of Edit Vector}
To analyze the properties of edit vectors, we analyze the 3,859 edits in SEEDA's GPT-3.5 output.

\begin{table}[t]
    \centering
    \small
    \begin{tabular}{l|c}
    \toprule
    Error Type & $L^2$ Norm (Mean $_\text{± Stdev.}$)   \\
    \midrule
ORTH       & 0.170 $_\text{± 0.406}$ \\
PUNCT      & 0.835 $_\text{± 0.597}$ \\
ADV        & 0.944 $_\text{± 0.485}$ \\
ADJ        & 0.962 $_\text{± 0.702}$ \\
NOUN:NUM   & 0.977 $_\text{± 0.593}$ \\
MORPH      & 0.977 $_\text{± 0.577}$ \\
PREP       & 1.033 $_\text{± 0.477}$ \\
VERB:FORM  & 1.066 $_\text{± 0.473}$ \\
DET        & 1.084 $_\text{± 0.558}$ \\
VERB:TENSE & 1.087 $_\text{± 0.510}$ \\
NOUN       & 1.109 $_\text{± 0.633}$ \\
VERB       & 1.171 $_\text{± 0.666}$ \\
VERB:SVA   & 1.231 $_\text{± 0.654}$ \\
PRON       & 1.247 $_\text{± 0.683}$ \\
CONJ       & 1.247 $_\text{± 0.556}$ \\
OTHER      & 1.318 $_\text{± 0.685}$ \\
SPELL      & 1.409 $_\text{± 0.610}$ \\
    \bottomrule
    \end{tabular}
    \caption{The mean and standard deviation of the $L^2$ norm of edit vectors for each error type.}
    \label{tab:edit-vector-norm}
\end{table}

\begin{figure}[t]
\centering
\includegraphics[width=0.48\textwidth]{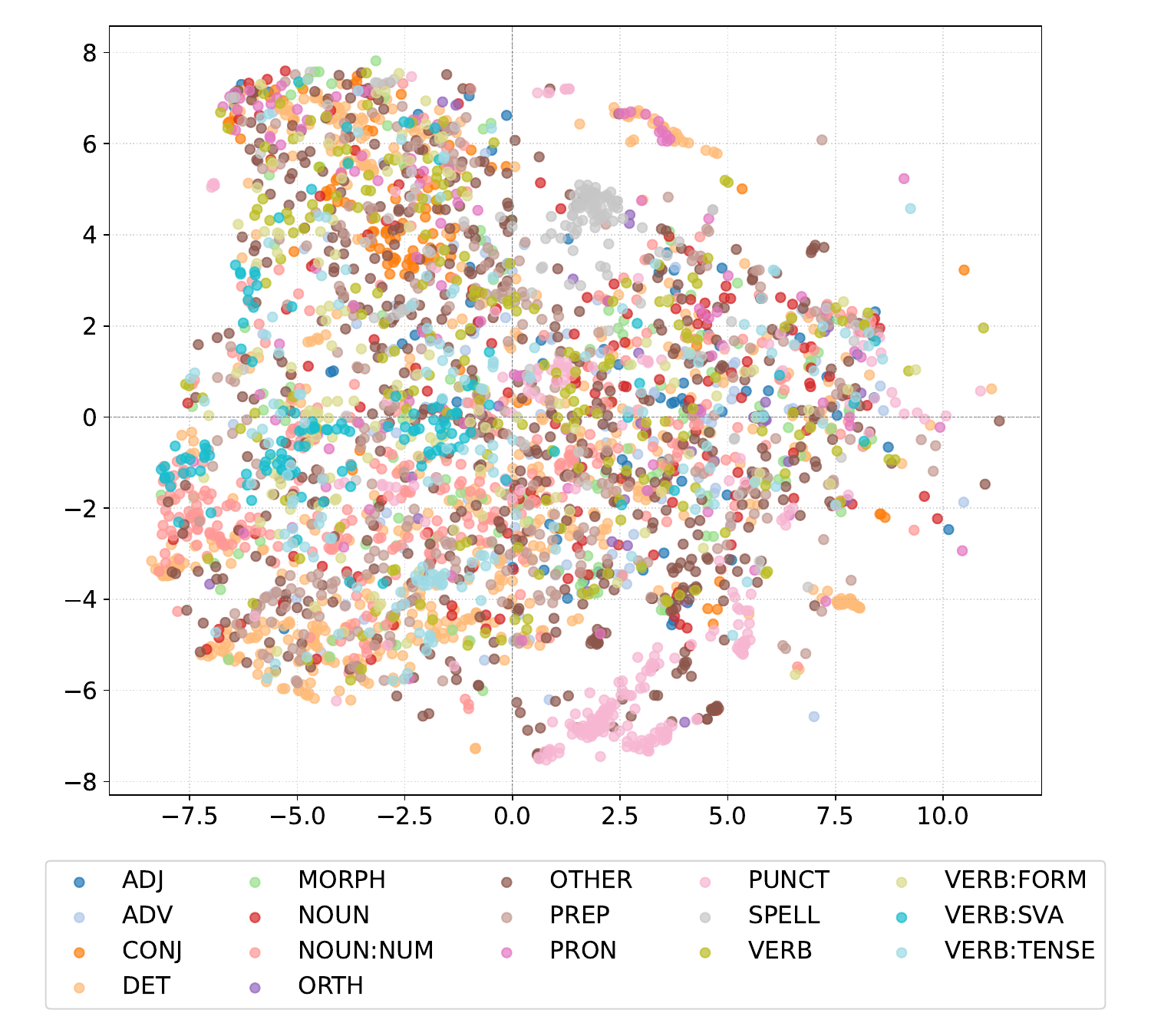}
\caption{Visualization of edit vectors with dimensionality reduced by t-SNE. The plot is color-coded according to the error type.}
\label{fig:edit-vector-vis}
\end{figure}

\subsubsection{Norm Varies Depending on Error Types}
The norm of the edit vector represents how strongly the edit changes the semantics of a sentence.
To investigate the relationship between the qualitative properties of edits and their norms, we focus on the error type of the edits. The error type is a classification system that describes the nature of the edit, primarily based on parts of speech. Examples of these classifications include SPELL, NOUN:NUM (Noun number), and DET (Determiner). Here, we use the error types proposed by ERRANT~\cite{bryant-etal-2017-automatic}.

Table~\ref{tab:edit-vector-norm} shows the mean and standard deviation of the edit vector norms for error types with a frequency of 50 or more. It is clear that the norm varies significantly depending on the error type. For instance, edits related to ORTH (Orthography, e.g., case and/or whitespace) and PUNCT (Punctuation) have smaller norms, while edits concerning content words such as VERB and NOUN have larger norms. 
Since edits to content words cause obvious changes in meaning, this tendency coordinates with our intuition.
In the context of UOT-ERRANT, the norm acts as an implicit weight for the edits, which have a direct contribution to the improvement in evaluation performance.

\begin{figure*}[t]
\centering
\begin{minipage}[b]{0.32\textwidth}
    \includegraphics[width=0.99\textwidth]{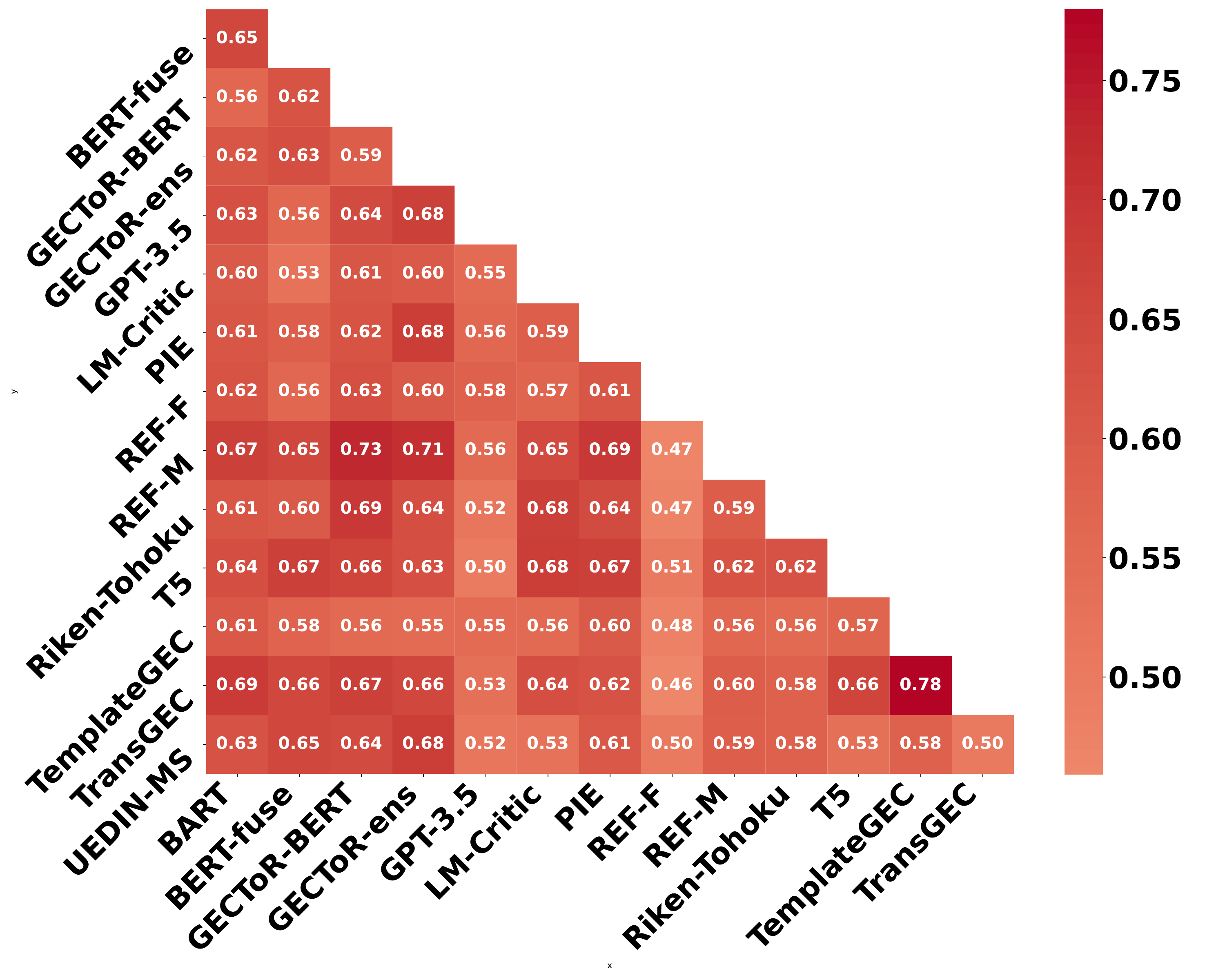}
    \subcaption{ERRANT}
    \label{fig:pairwise-errant}

\end{minipage}
  \begin{minipage}[b]{0.32\textwidth}
    \centering
    \includegraphics[width=0.99\textwidth]{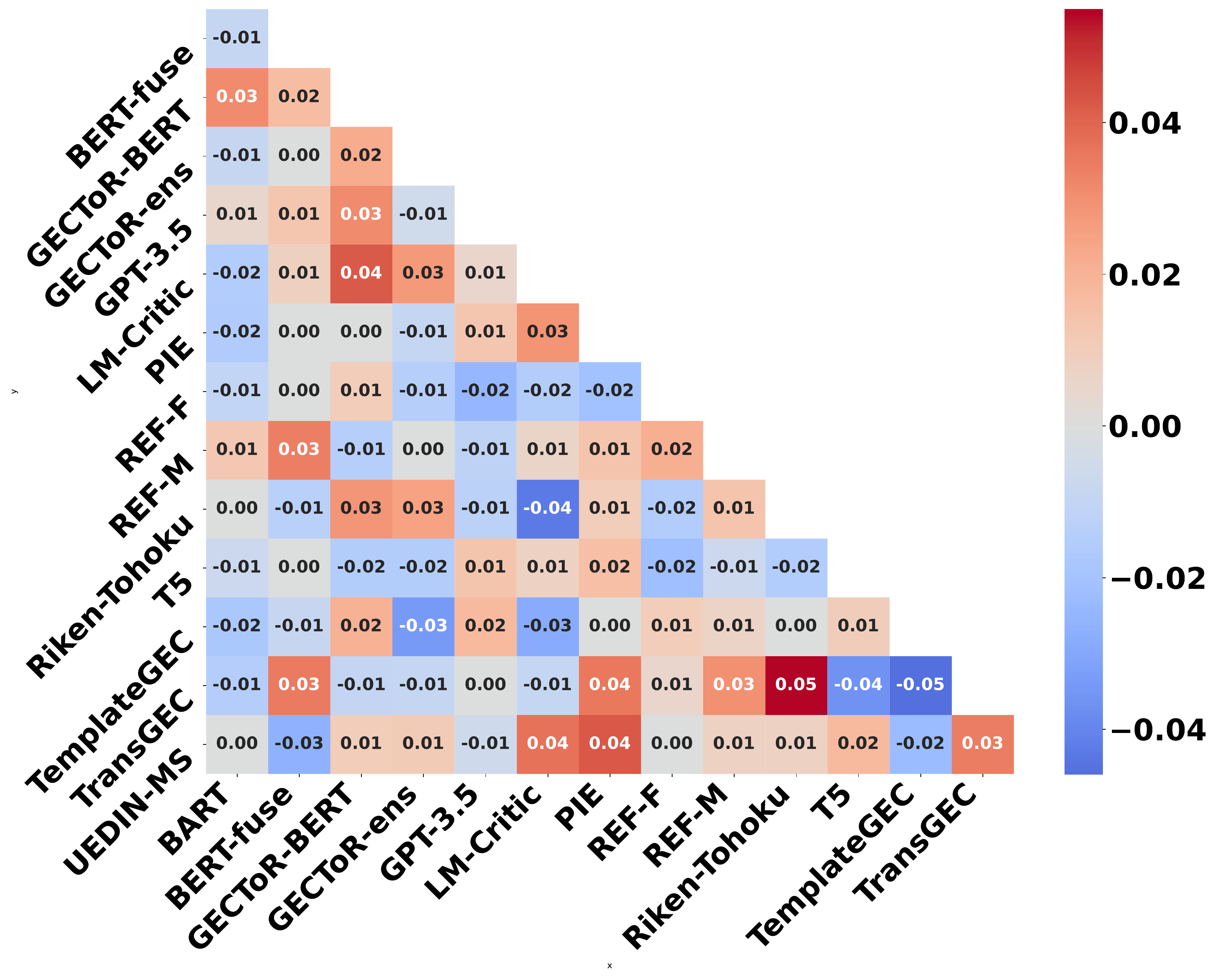}
    \subcaption{Improvement of PT-ERRANT}
    \label{fig:pairwise-diff-pterrant-errant}
\end{minipage}
\begin{minipage}[b]{0.32\textwidth}
    \includegraphics[width=0.99\textwidth]{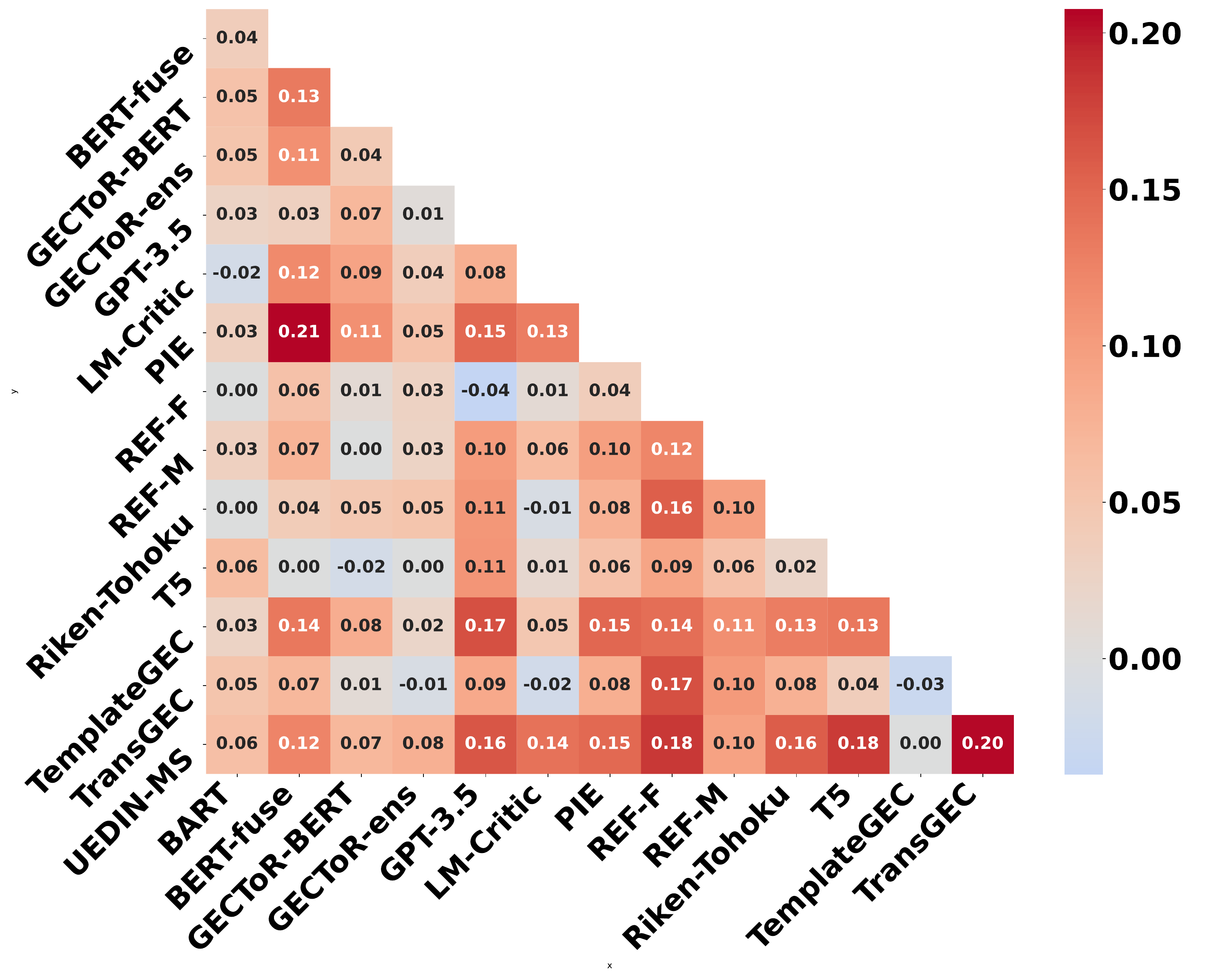}
    \subcaption{Improvement of UOT-ERRANT}
    \label{fig:pairwise-diff-uoterrant-errant}
\end{minipage}
  \caption{Sentence-level meta-evaluation results broken down by system pairs. Each cell shows the agreement rate between human and automatic evaluation in pairwise comparisons. ERRANT shows the absolute values, while PT-ERRANT and UOT-ERRANT indicate the improvement over ERRANT.
}
  \label{fig:pairwise}
\end{figure*}

\subsubsection{Similar Edits Become Similar Vectors}

To verify that similar edits are encoded in similar spaces, we use t-SNE~\cite{maaten2008visualizing} to reduce dimensionality and visualize the vectors for each error type. Figure~\ref{fig:edit-vector-vis} shows the plot for error types with a frequency of 50 or more. We observed that edits for PUNCT, SPELL, and CONJ (Conjunction) tend to form clusters. Furthermore, NOUN:NUM and VERB:SVA (Subject-verb agreement) are also positioned in proximity in the global view. As these two error types often share the commonality of adding an ``s'' to the end of a word, this suggests that the edit vector partially considers surface-level changes.

In contrast, edits such as NOUN, VERB, and ADJ (Adjective) are encoded in a scattered space. For these edits, the direction of semantic change varies depending on the vocabulary, thus the vectors within the error type group are not consistent. This result rather emphasizes that the edit vectors appropriately capture semantic changes.

\subsection{Accuracy of Pairwise Comparison}

Section~\ref{subsec:results} showed that UOT-ERRANT improves the correlation in +Fluency by assigning higher credits to systems such as GPT-3.5 and REF-F. Given that we convert sentence-level pairwise comparison scores into system rankings using TrueSkill, this improvement comes from better pairwise comparisons. To analyze this, we extend sentence-level meta-evaluation in GEC~\cite{kobayashi-etal-2024-revisiting}, which calculates the agreement rate of pairwise comparison results between human and automatic evaluations, to compute the agreement rate for each system pair.

Figure~\ref{fig:pairwise} illustrates the agreement rate of pairwise comparison results, broken down by system pair. We use +Fluency of SEEDA-E and NE-fluency as references. ERRANT's results (Fig.~\ref{fig:pairwise-errant}) are presented as absolute scores, while the agreement rates for PT-ERRANT (Fig.~\ref{fig:pairwise-diff-pterrant-errant}) and UOT-ERRANT (Fig.~\ref{fig:pairwise-diff-uoterrant-errant}) are shown as improvement from ERRANT. For ERRANT, the agreement rate tends to drop when pairs include GPT-3.5 or REF-F. Although PT-ERRANT extends ERRANT by adding edit weighting, it shows no meaningful improvement in the agreement rates. In contrast, UOT-ERRANT improves the agreement rate for pairs involving GPT-3.5 or REF-F by over 10\% in some cases. This improvement in pairwise comparison directly contributes to the higher correlation observed in Section~\ref{subsec:results}.

\subsection{\update{Computation Cost}}\label{subsec:efficiency}

\begin{figure}[t]
\centering
\includegraphics[width=0.48\textwidth]{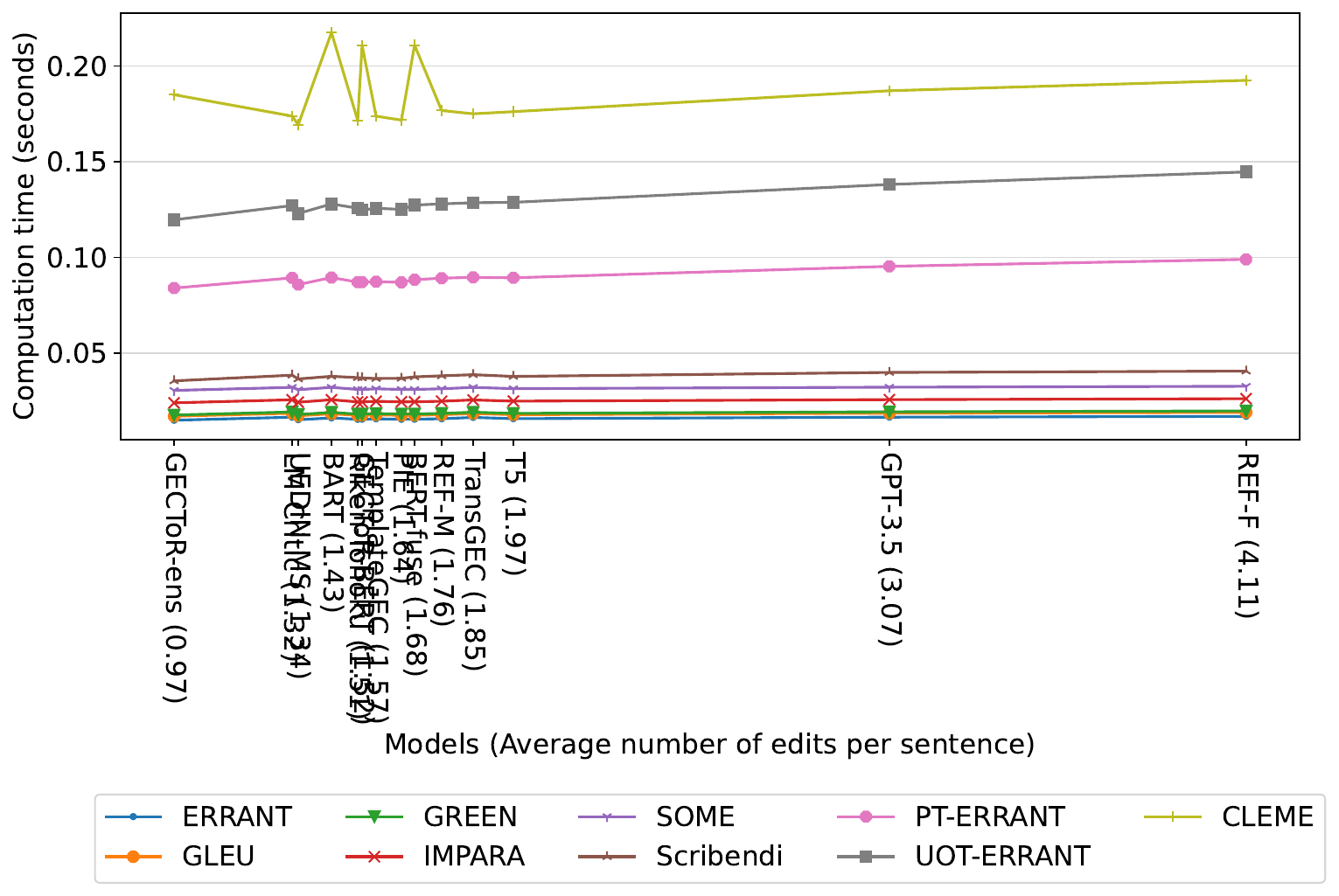}
\caption{\update{Actual average computation time per sentence (seconds). The horizontal axis shows system names and the average number of edits per sentence. The systems are plotted with intervals corresponding to the number of edits per sentence which is shown in the parentheses.}}
\label{fig:efficiency}
\end{figure}
\update{
Since the proposed method uses a neural model to calculate edit vectors, its computational cost is higher than that of ERRANT. While this could be a weakness, it is a necessary cost given the interpretability of UOT-ERRANT and the high correlation observed in the +Fluency setting of SEEDA-E. To experimentally investigate the actual computational cost, we measure the time required to evaluate each of the 13 systems used in the +Fluency setting of SEEDA. Each output has 391 sentences, but the number of edits is different across systems. In particular, we focus on the computation time per system to examine how the cost varies relative to the number of edits. This is also intended to verify whether larger-scale experiments will remain feasible when dealing with domains that contain a higher number of edits in the future. We use the metrics used in Table~\ref{tab:system-results}, but we exclude LLM-S and LLM-E because they scores multiple system outputs at once, thus we cannot compute system-wise computation time. All experiments were conducted on a single NVIDIA RTX 3090.
}

\update{
The results are shown in Figure~\ref{fig:efficiency}. UOT-ERRANT tends to have a higher computational cost compared to other metrics. First, compared to ERRANT, GLEU, and GREEN, UOT-ERRANT demands approximately eight times more time because it requires neural model computations. Second, while SOME, IMPARA, and Scribendi also require neural model computations, they are sentence-level metrics, meaning that the number of forward passes equal to the number of sentences. In contrast, UOT-ERRANT requires a number of passes proportional to the number of edits. Actually, the computation time tends to increase when evaluating systems with a large number of edits, such as GPT-3.5 or REF-F. Next, when compared with PT-ERRANT, UOT-ERRANT took about 1.5 times longer. This difference in cost can be explained by the presence of optimal transport calculation. Finally, CLEME exhibited the highest cost due to the implementation problem: inclusion of file I/O from executing multiple scripts in the CLI. Other metrics do not encounter this problem as they can be run in a single script via \textsc{gec-metrics}~\cite{goto-etal-2025-gec}. Although these are implementation-level differences, the same issues would arise in actual use cases.
}

\update{
In summary, while UOT-ERRANT requires a relatively high computational cost, we believe it is acceptable considering the improvement in evaluation performance shown in Table~\ref{tab:system-results}. Since evaluation is fundamentally a one-time process, the computational cost is not a major problem. It might become an issue in the future if online execution is required, such as when applying it as a reward for reinforcement learning. A potential solution is to accelerate the computation by caching the results of the embedding model or by implementing appropriate batching. For example, if multiple GEC systems output the same corrected sentence, the result for one can be reused for the others. 
}

\section{Conclusion}
In this study, we propose a new GEC metric based on \emph{edit vectors} to calculate the similarity between a hypothesis and a reference edits. The edit vector is defined as the difference vector between sentence representations before and after removing edits, providing a novel way to quantify edits. Furthermore, we propose UOT-ERRANT, based on the idea of optimally transporting edit vectors from the hypothesis to the reference. Meta-evaluation using the SEEDA dataset showed that UOT-ERRANT improved conventional edit-level metrics, particularly in domains with a high number of edits. Our analysis revealed that edit vectors effectively reflect the qualitative characteristics of edits, demonstrated by analyzing their norms and clusters based on error types.

Our proposed \emph{edit vector} can be applied to a variety of tasks involving partial edits. For example, in text simplification~\cite{shardlow-etal-2024-bea}, rewriting a difficult word or phrase into a simpler one can be regarded as an edit. Similarly, in automatic speech recognition error correction~\cite{leng-etal-2021-fastcorrect-2}, edits occur much like in GEC, allowing for the calculation of difference vectors. In the vision field, image editing tasks are well-known~\cite{avrahami2022blended}. Images before and after editing can be encoded separately by an image encoder, and their difference can be taken to vectorize the edit. In the future, we anticipate that applying the edit vector to other tasks will broaden its applications, particularly in evaluation-centric scenarios.
The limitations and ethical statements of this study are discussed in Appendices~\ref{sec:limitation} and~\ref{sec:ethic}.

\section*{Acknowledgments}
We thank the action editor and the anonymous reviewers for their valuable comments. This work has been supported by JST SPRING Grant Number JPMJSP2140.

\bibliography{tacl2021}
\bibliographystyle{acl_natbib}

\clearpage
\appendix

\section*{Appendix}

\section{Limitations}\label{sec:limitation}
\paragraph{Meta-evaluation}
This paper conducted meta-evaluation using SEEDA and confirms an improved correlation with edit-level human evaluation. This meta-evaluation includes both Base and +Fluency domains, assumed to cover general use cases. However, for other cases, it's also possible to consider domains with low error density, as proposed in CWEB~\cite{flachs-etal-2020-grammatical}, or the correction of texts by writers not limited to learners, as discussed in GMEG-data~\cite{napoles-etal-2019-enabling}. Current meta-evaluation relying on SEEDA has limitations in such domains, but we look forward to future extensions of meta-evaluation datasets.

\paragraph{Encoder Models}
Since edit vectors are based on sentence representations, the quality of the vectors depends on the quality of the encoder model. In this study, we adopted ELECTRA because its pre-training tasks are similar to grammatical error detection, but other models might show better evaluation performance. 
As shown in Section~\ref{tab:ablation}, the optimal encoder varies depending on the target domain of evaluation. While we could explore the optimal encoder in this study, it would merely reproduce overfitting to the specific benchmark. Given that users practically need to find the best model for their specific domain, we believe there is little value in discussing encoder selection in depth here.
\update{Furthermore, it is also interesting to investigate which model characteristics lead to higher-quality editing vectors. While model probing may be useful for this purpose, the methodology for investigating its relationship with editing vectors has not yet been fully established and remains it for future work.}

\paragraph{\update{Computation Cost}}
\update{
As stated in Section~\ref{subsec:efficiency}, the proposed method requires more computation time than other metrics because it involves neural model computations for each edit. However, as shown in Figure~\ref{fig:efficiency}, the computation time per sentence is less than 0.15 seconds, which we consider to be sufficiently fast. Furthermore, since a single execution is typically sufficient for evaluation, the computational cost does not become a major problem in most cases.
}

\section{Ethical Statements}\label{sec:ethic}
\paragraph{Social Bias}
The proposed method implicitly weights edits by using the norm of the edit vectors. As shown in the ablation study (\S \ref{subsec:ablation}), while this weighting is important, it can also lead to bias. For example, if the weight of an edit changes based on different social attributes like gender or nationality, there's a risk that the metric could favor certain attributes. However, this issue is not exclusive to UOT-ERRANT; it could also exist in PT-ERRANT and LLM-based metrics. We would like to defer a comprehensive discussion on this point to future research.

\end{document}